\documentclass[twocolumn,10pt]{asme2e}

\newcommand{\be}{\begin{equation}}
\newcommand{\ee}{\end{equation}}
\newcommand{\beq}{\begin{equation}}
\newcommand{\eeq}{\end{equation}}
\newcommand{\bed}{\begin{displaymath}}
\newcommand{\eed}{\end{displaymath}}
\newcommand{\beqa}{\begin{eqnarray}}
\newcommand{\eeqa}{\end{eqnarray}}
\newcommand{\beqann}{\begin{eqnarray*}}
\newcommand{\eeqann}{\end{eqnarray*}}
\newcommand{\bseq}{\begin{subequations}}
\newcommand{\eseq}{\end{subequations}}

\newcommand{\ba}{\begin{array}}
\newcommand{\ea}{\end{array}}

\newcommand{\1}{{\bf 1}}

\usepackage{amsbsy}
\usepackage[dvips]{graphicx}
\usepackage{psfrag}
\usepackage{epsf}
\usepackage{epsfig}
\usepackage{array}
\usepackage{amssymb}
\usepackage{subfigure}
\usepackage{subeqn}

\confshortname{DETC'2004}
\conffullname{ASME Design Engineering Technical Conferences and \\ \raggedleft Computers and Information in Engineering Conferences}
\confdate{}
\confmonth{September 28 -October 2}
\confyear{2004}
\confcity{Salt Lake City, Utah}
\confcountry{USA}

\papernum{DETC2004-57682}

\title{The Virtual Manufacturing concept: Scope, Socio-Economic Aspects and Future Trends. }
\author{P. D\'epinc\'e, D. Chablat
    \affiliation{
      Institut de Recherche en Communications \\et Cybern\'etique de
      Nantes
      \thanks{IRCCyN: UMR n$^\circ$ 6597 CNRS, \'Ecole Centrale de Nantes,
                        Universit\'e de Nantes, \'Ecole des Mines de
                        Nantes}\\
      1, rue de la No\"e, 44321 Nantes, France \\
      Philippe.Depince@irccyn.ec-nantes.fr
    }}
\author{E. No\"{e}l
    \affiliation{
      CECIMO\\
      66 avenue Louise, \\1050 Brussels, Belgium \\
      eco-dep@cecimo.be
    }}
\author{P.O. Woelk
    \affiliation{
    Institute of Production Engineering \\and Machine Tools (IFW)\\
    Schlosswender Strasse 5,\\ D-30159 Hannover, Germany \\
    Woelk@ifw.uni-hannover.de
    }}
\begin{document}
\maketitle
\begin{abstract}
The research area ``Virtual Manufacturing (VM)'' is the use of information technology and computer simulation to model real world manufacturing processes for the purpose of analysing and understanding them. As automation technologies such as CAD/CAM have substantially shortened the time required to design products, Virtual Manufacturing will have a similar effect on the manufacturing phase thanks to the modelling, simulation and optimisation of the product and the processes involved in its fabrication. After a description of Virtual Manufacturing (definitions and scope), we present some socio-economic factors of VM and finaly some ``hot topics'' for the future are proposed.

\end{abstract}
\section*{INTRODUCTION}
Manufacturing is an indispensable part of the economy and is the
central activity that encompasses product, process, resources and
plant. Nowadays products are more and more complex, processes are
highly-sophisticated and use micro-technology and mechatronic, the
market demand (lot sizes) evolves rapidly so that we need a
flexible and agile production. Moreover manufacturing enterprises
may be widely distributed geographically and linked conceptually
in terms of dependencies and material, information and knowledge
flows. In this complex and evolutive environment, industrialists
must know about their processes before trying them in order to get
it right the first time. To achieve this goal, the use of a
virtual manufacturing environment will provide a computer-based
environment to simulate individual manufacturing processes and the
total manufacturing enterprise. Virtual Manufacturing systems
enable early optimization of cost, quality and time drivers,
achieve integrated product, process and resource design and
finally achieve early consideration of producibility and
affordability. The aim of this paper is to present an updated
vision of Virtual Manufacturing (VM) through different aspects. This vision is
the result of a survey done within the thematic European Network MANTYS.
As, since 10 years, several projects and workshops have dealt with
the Virtual Manufacturing thematic, we will first define the
objectives and the scope of VM and the domains that are concerned.
The expected technological benefits of VM will also been
presented. In a second part, we will present the socio-economic
aspects of VM. This study will take into account the market
penetration of several tools with respect to their maturity, the
difference in term of effort and level of detail between
industrial tools and academic research. Finally the expected
economic benefits of VM will be presented and a focus will be made on SMEs.
 The last part will describe the trends and exploitable results in machine tool
industry (research and development towards the ``Virtual Machine
Tool''), automotive (Digital Product Creation Process to design
the product and the manufacturing process) and aerospace.
A brief description of ``hot topics'' is proposed.

\section*{MANTYS: THEMATIC NETWORK ON MANUFACTURING TECHNOLOGIES}
The MANTYS Thematic Network is supported by the European
Commission Growth Programme (FP5) and promotes innovation in the
field of manufacturing technologies, focusing on machinery. It
provides a European platform that enables research and industrial
participants to exchange views and research results in technology,
socio-economic issues, sustainability and the quality of life in
all aspects of manufacturing. It was launched in September 2001
and brings together:
\begin{itemize}
\item -over 20 European RTD laboratories and research institutes with eminent reputations in manufacturing technologies,
\item -a socio-economic task force composed of various experts from universities, private companies and machine tool builders' associations,
\item -an Industrial Advisory Committee representing the automotive, aerospace, mechanical and machine building sectors.
\end{itemize}
The MANTYS Thematic Network has established a framework in which
researchers and industrialists communicate and reach common view
of technological progress and its opportunities. Potential
synergies between different projects are made apparent and
co-operation promoted, to enhance the creation of the European
Research Area in manufacturing technologies. MANTYS was actually
proved to be a very efficient communication platform at the launch
of the 6th Framework Programme, by allowing researchers and EC
officials to meet one another and discuss proposals for FP6
projects.

Furthermore, the MANTYS Thematic Network promotes the widest
exploitation of research results across sectors. Results from over
80 research projects have been so far assessed and used to create
Technology Trend Reports on 5 key technological areas:
Reliability, Process monitoring -  Control; Agility - Flexibility;
Process ECO-Efficiency; Precision Engineering and Virtual
Manufacturing.

MANTYS is generating insight into the mechanisms that relate
technological innovation to socio-economic factors. Building on
this, it will identify realistic scenarios based on likely
technological and socio-economical developments. These will be
used to assess strategic impact. Also, ``Technology Road Maps''
will be produced, preparing decision makers to orient their
activities and to adapt to change.
\subsection*{Virtual Manufacturing technology area}
\begin{figure}[!ht]
  \center
  \epsfig{file=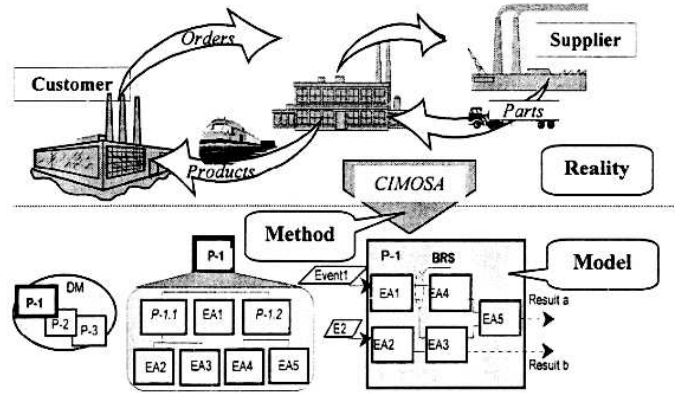, width=9cm}
  \caption{ENTERPRISE MODELING - CIMOSA MODEL}
  \label{CIMOSA}
 \end{figure}

Virtual Manufactturing has always been supported by the European
commission during all the framework programs. During the FP3, the
CIM-OSA association has been creating and it is stil active. This
European organization is attempting to develop Open System
Architecture (OSA) standards for CIM (Computer Integrated
Manufacturing) environments. CIMOSA \cite{kosanke} provides a
process oriented modelling concept that captures both the process
functionality and the process behaviour. It supports evolutionary
enterprise modelling, e.g., the modelling of individual enterprise
domains (DM) which may contain one or several individual processes
(P-1, P-2, ...)  (Fig.~\ref{CIMOSA}).

The aims of MANTYS in the field of Virtual Manufacturing are to
promote the successful exploitation of innovative technologies
across sectors, establish a continuously updated view on
technological trends and identify strategies and scenarios for
future technological Research and Development. The tools used to
achieve this objectives are : the establishment of innovation
sheets on different European or national project dealing with
Virtual Manufacturing - some examples are given in Table
\ref{Table2} -, the establishment of trend reports and public
information dissemination.

 \begin{table}[!ht]
\center
 {\small
 \begin{tabular}{|m{1.5cm}|m{6.5cm}|}
 \hline
\textbf{Acronym} &
 \textbf{Short description} \\ \hline
Fashion &
 Fully integrated simulation and optimization for enhancing advanced machine tools design \\ \hline
 MECOMAT &
Mechatronic compiler for machine tool design\\ \hline
MICROHARD &
Development of hard turning towards a micron accuracy capability process for serial production\\ \hline
ProREAL&
Virtual reality based methods and tools for applications related with planning and training tasks in manufacturing processes\\ \hline
RRS&
Realistic Robot Simulation\\ \hline
SAM&
New structural alternative for modules in machinery\\ \hline
SPI-9&
Design System for simultaneous engineering applied to production systems \\ \hline
VIR-ENG &
Integrated design, simulation and distributed control of agile modular manufacturing machinery \\ \hline
... &
...\\ \hline
 \end{tabular}}
 \caption{EC FP5 projects linked to VM}
 \label{Table2}
 \end{table}
\section*{WHAT IS VIRTUAL MANUFACTURING?}

\subsection*{Virtual manufacturing definitions}
\begin{figure}[!ht]
\center
  \epsfig{file=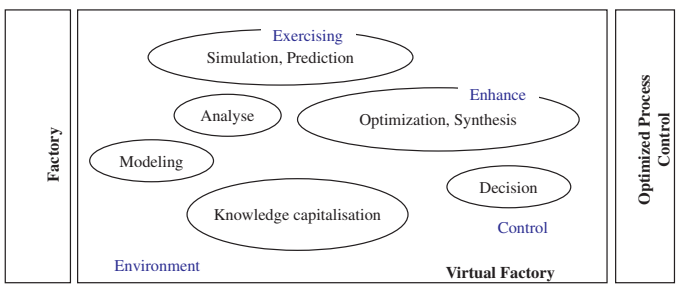, width=9cm}
  \caption{VIRTUAL MANUFACTURING}
  \label{Fig1}
 \end{figure}

The term Virtual Manufacturing is now widespread in literature but
several definitions are attached to these words. First we have to
define the objects that are studied. Virtual manufacturing
concepts originate from machining operations and evolve in this
manufacturing area. However one can now find a lot of applications
in different fields such as casting, forging, sheet metalworking
and robotics (mechanisms). The general idea one can find behind
most definitions is that ``Virtual Manufacturing is nothing but
manufacturing in the computer''. This short definition comprises
two important notions: the process (manufacturing) and the
environment (computer). In \cite{iwata,lee} VM is defined as
``...manufacture of virtual products defined as an aggregation of
computer-based information that... provide a representation of the
properties and behaviors of an actualized product''. Some
researchers present VM with respect to virtual reality (VR). On
one hand, in \cite{bowyer} VM is represented as a virtual world
for manufacturing, on the other hand, one can consider virtual
reality as a tool which offers visualization capabilities for VM
\cite{lin,Cobb}. The most comprehensive definition has been
proposed by the Institute for Systems Research, University of
Maryland, and discussed in \cite{lawrence,saadoun}. Virtual
Manufacturing is defined as ``an integrated, synthetic
manufacturing environment exercised to enhance all levels of
decision and control'' (Fig.~\ref{Fig1}).

{\bf Environment:} supports the construction, provides tools,
models, equipment, methodologies and organizational principles,

{\bf Exercising:} constructing and executing specific
manufacturing simulations using the environment which can be
composed of real and simulated objects, activities and processes,

{\bf Enhance:} increase the value, accuracy, validity,

{\bf Levels:} from product concept to disposal, from factory
equipment to the enterprise and beyond, from material
transformation to knowledge transformation,

{\bf Decision:} understand the impact of change (visualize,
organize, identify alternatives).

A similar definition has been proposed in [8]: ``Virtual
Manufacturing is a system, in which the abstract prototypes of
manufacturing objects, processes, activities, and principles
evolve in a computer-based environment to enhance one or more
attributes of the manufacturing process.''

One can also define VM focusing on available methods and tools
that allow a continuous, experimental depiction of production
processes and equipment using digital models. Areas that are
concerned are (i) product and process design, (ii) process and
production planning, (iii) machine tools, robots and manufacturing
system and virtual reality applications in manufacturing.
\subsection*{The scope of Virtual Manufacturing}
The scope of VM can be to define the product, processes and
resources within cost, weight, investment, timing and quality
constraints in the context of the plant in a collaborative
environment. Three paradigms are proposed in \cite{lawrence}:
\begin{itemize}
\item -{\bf Design-centered VM:} provides manufacturing information to
the designer during the design phase. In this case VM is the use
of manufacturing-based simulations to optimize the design of
product and processes for a specific manufacturing goal (DFA,
quality, flexibility, ...) or the use of simulations of processes
to evaluate many production scenario at many levels of fidelity
and scope to inform design and production decisions.

\item -{\bf Production-centered VM:} uses the simulation capability to
modelize manufacturing processes with the purpose of allowing
inexpensive, fast evaluation of many processing alternatives. From
this point of view VM is the production based converse of
Integrated Product Process Development (IPPD) which optimizes
manufacturing processes and adds analytical production simulation
to other integration and analysis technologies to allow high
confidence validation of new processes and paradigms.

\item  -{\bf Control-centered VM:} is the addition of simulations to control
models and actual processes allowing for seamless simulation for
optimization during the actual production cycle.
\end{itemize}
Another vision is proposed by Marinov in \cite{marinov1,marinov2}. The activities in
manufacturing include design, material selection, planning,
production, quality assurance, management, marketing, .... If the
scope takes into account all these activities, we can consider
this system as a Virtual Production System. A VM System includes
only the part of the activities which leads to a change of the
product attributes (geometrical or physical characteristics,
mechanical properties, ...) and/or processes attributes (quality,
cost, agility, ...). Then, the scope is viewed in two directions:
a horizontal scope along the manufacturing cycle, which involves
two phases, design and production phases, and a vertical scope
across the enterprise hierarchy. Within the manufacturing cycle,
the design includes the part and process design and, the
production phase includes part production and assembly.

\begin{figure}[!ht]
\center
  \epsfig{file=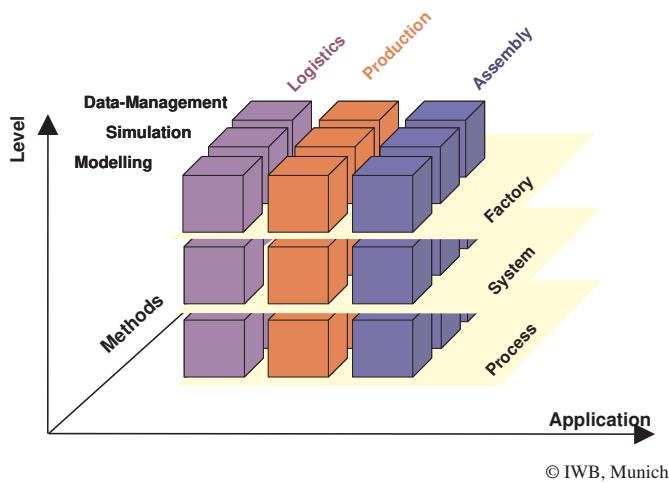, width=9cm}
  \caption{VIRTUAL MANUFACTURING OBJECTIVES, SCOPES AND DOMAINS}
  \label{Fig2}
 \end{figure}

We choose to define the objectives, scope and the domains
concerned by the Virtual Manufacturing thanks to the 3D matrix
represented in Fig.~\ref{Fig2}.

The vertical plans represent the three main aspects of
manufacturing today: Logistics, Productions and Assembly, which
cover all aspects directly related to the manufacturing of
industrial goods. The horizontal planes represent the different
levels within the factory. At the lowest level (microscopic
level), VM has to deal with unit operations, which include the
behavior and properties of material, the models of machine tool -
cutting tool - workpiece - fixture system. These models are then
encapsulated to become VM cells inheriting the characteristics of
the lower level plus some extra characteristics from new objects
such as a virtual robot.
Finally, the macroscopic level (factory level) is derived from all
relevant sub-systems. The last axis deals with the methods we can
use to achieve VM systems. These methods will be discussed in the
next paragraph.
\subsection*{Methods and tools used in Virtual manufacturing}
Two main activities are at the core of VM. The first one is the
``modeling activity'' which includes determining what to model and
the degree of abstraction that is needed. The second one is the
ability to represent the model in a computer-based environment and
to correlate to the response of the real system with a certain
degree of accuracy and precision: the ``simulation activity''.
Even if simulation tools often appears to be the core activity in
VM, others research areas are relevant and necessary. One can find
in \cite{lawrence} a classification of the technologies within the context of
VM in 4 categories:

\begin{itemize}
\item  A ``Core'' technology is a technology which is
fundamental and critical to VM. The set of ``Core'' technologies
represents what VM can do.

\item An ``Enabling'' technology is necessary
to build a VM system.

\item A ``Show stopper'' technology is one without
which a VM system cannot be built.

\item  ``Common''
technology is one that is widely used and is important to VM.
\end{itemize}
We propose the following activities to underline methods that are
necessary to achieve a VM system:

-{\it Manufacturing characterization:} capture, measure and
analyze the variables that influence material transformation
during manufacturing (representation of product/process, design by
features, system behavior, ...),

-{\it Modeling and representation technologies:} provide different
kinds of models for representation, abstraction, standardization,
multi-use, ... All the technologies required to represent all the
types of information associated with the design and fabrication of
the products and the processes in such a way that the information
can be shared between all software applications (Knowledge based
systems, Object oriented, feature based models, ...)

-{\it Visualization, environment construction technologies:}
representation of information to the user in a way that is
meaningful and easily comprehensible. It includes Virtual reality
technologies, graphical user interfaces, multi context analysis
and presentation.

-{\it Verification, validation and measurement:} all the tools and
methodologies needed to support the verification and validation of
a virtual manufacturing system (metrics, decision tools, ...).

-{\it Multi discipline optimization:} VM and simulation are
usually no self-standing research disciplines, they often are used
in combination with ``traditional'' manufacturing research.

Numerous tools are nowadays available for simulating the different
levels described in Fig.~\ref{Fig2}: from the flow simulation
thanks to discrete event simulation software to finite elements
analysis. The results of these simulations enable companies to
optimise key factors which directly affects the profitability of
their manufactured products. Table \ref{Table1} proposes an overview of
simulation applications in manufacturing.

-{\it Flow simulation - discrete event simulation software}: It is composed of object-oriented discrete event simulation
tool to efficiently model, experiment and analyze facility layout and process flow. It aids at the determination of optimal layout, throughput,
cost and process flow for existing or new systems. It simulates and optimizes production lines in order to accommodate different order sizes and product mixes.

-{\it Graphical 3D - Kinematics simulation}: There are used in robotic and machine tools simulation tools
for design, evaluation, and off-line programming of workcells.
They can incorporate real world robotic and peripheral equipment, motion attributes, kinematics, dynamics and I/O logic.

-{\it Finite elements analysis}: A powerfull engineering design tool that has enabled
companies to simulate all kind of fabrication and testing in a more realistic manner.
It can be use in combination with optimisation tool as a tool for decision making.
It allows to reduce the number of prototypes as virtual prototype are cheaper than building physical models, the material waste and the cost of tooling,...

 \begin{table}
\center
 {\footnotesize
 \begin{tabular}{|m{1.5cm}|m{2.2cm}|m{2.5cm}|m{1.2cm}|}
 \hline
 Manufacturing Level &
 Type of simulation &
 Simulation targets  &
 Level of detail \\ \hline
 Factory / \goodbreak shop floor  &
 -Flow simulation \goodbreak -Business process \goodbreak simulation &
 -Logistic and storage \goodbreak -Production principles \goodbreak -Production planning and control&
 low  \\ \hline
 Manufacturing systems / \goodbreak manufacturing lines &
 -Flow simulation &
 -System layout \goodbreak-Material flow \goodbreak-Control strategies \goodbreak-System capacity \goodbreak-Personnel planning&
 Intermediate\\ \hline
 Manufacturing cell / \goodbreak Machine tool / robot &
 -Flow simulation \goodbreak -Graphical 3D \goodbreak kinematics simulation &
 -Cell layout \goodbreak -Programming \goodbreak-Collision test &
 High\\ \hline
 Components  &
 -Finite-Elements Analysis \goodbreak -Multibody simulation
 \goodbreak-Bloc simulation &
 -Structure (mechanical and thermal) \goodbreak-Electronic circuits \goodbreak -Non-linear movement dynamics &
 Complex \\ \hline
 Manufacturing processes &
 -Finite-Elements analysis  &
 -Cutting processes: surface properties, thermal effects, tool wear / life
time, chip creation \goodbreak -Metal forming processes: formfill,
material flow (sheet metal), stresses, cracks &
 Very \goodbreak complex \\ \hline
 \end{tabular}}
 \caption{ SIMULATIONS TOOLS}
 \label{Table1}
 \end{table}
\section*{ECONOMICS AND SOCIO-ECONOMICS FACTOR OF VIRTUAL MANUFACTURING}

\subsection*{Expected benefits}
As small modifications in manufacturing can have important effects
in terms of cost and quality, Virtual Manufacturing will provide
manufacturers with the confidence of knowing that they can deliver
quality products to market on time and within the initial budget.
The expected benefits of VM are:

-from the product point of view it will reduce time-to-market,
reduce the number of physical prototype models, improve quality,
...: in the design phase, VM adds manufacturing information in
order to allow simulation of many manufacturing alternatives: one
can optimize the design of product and processes for a specific
goal (assembly, lean operations, ...) or evaluate many production
scenarios at different levels of fidelity,

-from the production point of view it will reduce material waste,
reduce cost of tooling, improve the confidence in the process,
lower manufacturing cost,...: in the production phase, VM
optimizes manufacturing processes including the physics level and
can add analytical production simulation to other integration and
analysis technologies to allow high confidence validation of new
processes or paradigms. In terms of control, VM can simulate the
behavior of the machine tool including the tool and part
interaction (geometric and physical analysis), the NC controller
(motion analysis, look-ahead).

VM and simulation change the procedure of product and process
development. Prototyping will change to virtual prototyping so
that the first real prototype will be nearly ready for production.
This is intended to reduce time and cost for any industrial
product. Virtual manufacturing will contribute to the following
benefits \cite{lawrence}:

{\bf 1~Quality:} Design For Manufacturing and higher quality of
the tools and work instructions available to support production;

{\bf 2~Shorter cycle time:} increase the ability to go directly
into production without false starts;

{\bf 3~Producibility:} Optimise the design of the manufacturing
system in coordination with the product design; first article
production that is trouble-free, high quality, involves no reworks
and meets requirements.

{\bf 4~Flexibility:} Execute product changeovers rapidly, mix
production of different products, return to producing previously
shelved products;

{\bf 5~Responsiveness:} respond to customer ``what-ifs'' about the
impact of various funding profiles and delivery schedule with
improved accuracy and timeless,

{\bf 6~Customer relations:} improved relations through the
increased participation of the customer in the Integrated Product
Process Development process.
\subsection*{Economic aspects}
\begin{figure}[!ht]
  \center
  \epsfig{file=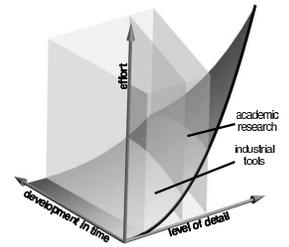, width=9cm}
  \caption{ACADEMIC RESEARCH VERSUS INDUSTRIAL TOOLS}
  \label{Fig3}
 \end{figure}

It is important to understand the difference between academic
research and industrial tools in term of economic aspects. The
shape of the face in the diagram presented in Fig.~\ref{Fig3} \cite{depince},
is defined by two curves:

-``effort against level of detail'' where ``level of detail''
refers to the accuracy of the model of simulation (the number of
elements in the mesh of a FEM model or the fact if only static
forces are taken into account for a simulation,

-``effort against development in time'' is a type of time axis and
refers to future progress and technological developments (e.g.
more powerful computers or improved VR equipment).

\begin{figure}[!ht]
\center
  \epsfig{file=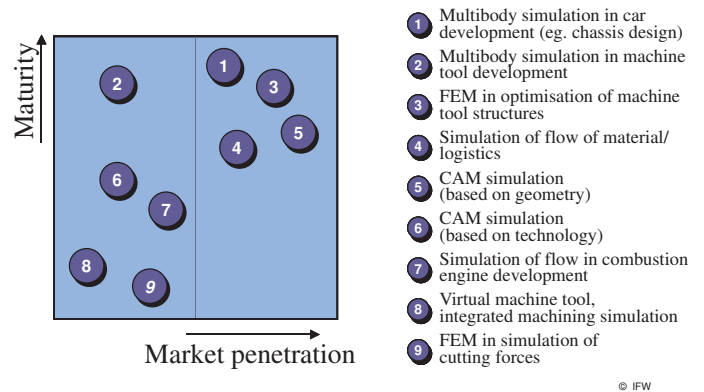, width=9cm}
  \caption{MATURITY OF TECHNIQUES VERSUS MARKET PENETRATION}
  \label{Fig4}
 \end{figure}

Universities develop new technologies focusing on technology
itself. Researchers do not care how long the simulation will need
to calculate the results and they not only develop the simulation
but they need to develop the tools and methods to evaluate whether
the simulation is working fine and whether the results are exact.
On the other hand, industrial users focus on reliability of the
technology, maturity economic aspects (referring to the effort
axis) and on the integration of these techniques within existing
information technology systems of the companies (e.g. existing
CAD-CAM systems, ...). To our mind, Virtual Manufacturing is, for
a part of its scope, still an academic topic. But in the future,
it will become easier to use these technologies and it will move
in the area of industrial application and then investments will
pay off. For example in the automotive and aerospace companies in
the late 60's, CAD was struggling for acceptance. Now 3-D geometry
is the basis of the design process. It took 35 years for CAD-CAM
to evolve from a novel approach used by pioneers to an established
way of doing things. During this period, hardware, software,
operating systems have evolved as well as education and
organizations within the enterprise in order to support these new
tools. Fig.~\ref{Fig4} presents a classification of VM
applications with respect to market penetration and maturity.
Three classes can be determined: some techniques are daily used in
industry, some are mature but their uses are not widespread and
some are still under development.

No doubt that the current developments in virtual manufacturing
and the growing interest of the industries for this area are
influenced by the changes that have recently taken place in our
society. The old industrial model, which was characterized by
``real'' activities, work and capital, is over. Focusing on
manufacturing technology domain, four stages in its evolution can
be identified \cite{Offodile}:

-{\it Craftsman (pre-1870)}: Highly flexible system with economy
of scope and quality. Mostly artisans who started from the design
until the delivery,

-{\it mass production (1870-1945)}: Standardization,
interchangeability of parts. Mechanization of the manufacturing
process. Economy of scale justified the high cost of the
production machines,

-{\it automation technology (1945-1995)}: Automation served to bridge the gap between the highly flexible \textit{craftsman}
and the highly volume dependent \textit{mass production},

-{\it IT-based manufacturing network (1995-present)}: CNCs, robots, FMS, ... often linked via local area networks.
The emergence of IT and Internet enables the development of new distributed networks.

Today, the work has been redistributed. Shareholders run companies
and evaluate activities globally. In this new frame, Virtual
Manufacturing will allow companies to maximise the shareholder
value by enabling companies that operate in a global market to
take advantage of the globalisation of exchanges and economic
activities. It will be easier to transfer mass production to a
specific country corresponding to certain criteria, while others
are developing the ideas. IT will reduce organizational and
geographical barriers to collaboration and can replace physical
movement (of people, materials, ...) with digital movement.
However question dealing with the balancing of these new
activities remains open and leads to various ethical questions.

One of the major advantages of Virtual Manufacturing for large
companies is the possibility to set up virtual organisations. With
such a tool, companies can now concentrate on core business and
outsource all secondary activities. The main reason for companies
to transfer and outsource activities is that it increases the cost
efficiency and release resources for more essential projects. The
development of the know-how and interfaces as well as security
aspects of a Virtual Manufacturing system should not be
underestimated and remain critical in the decision process to
invest or not in Virtual Manufacturing.

\subsection*{SMEs and Virtual Manufacturing}
As presented earlier, VM appeared to be a set of IT technologies
which are now mature and whose benefits for industry are numerous.
If large manufacturing enterprises (LMEs) have developed and
applied with success these technologies (automotive, aerospace,
...) it is not the case for the majority of SMEs. Publications
dealing with VM are mainly about new capabilities or products or
describe succes stories how they have been implemented in LMEs.

First of all, VM is a capital intensive technology and a lot of
SMEs do not have the wherewithal to integrate them. VM
technologies are still quite expensive to own. There are clear
differences between SMEs and LMEs \cite{Offodile}: the both of
them need growth and operate in a highly competitive market.
However, SMEs lack the basis competitive infrastructure such as
cash flow, trained employees, ... They are \textit{resource poor}
and they do not have time and money to speculate on IT.

As pointed out by recent reports, SMEs are the ``base'' of most
manufacturing economies and as manufacturing ``creates the real
wealth'' in any economy, SMEs can be seen as the backbone of these
economies. Due to this economic importance, it seems important to
have positive and sustained policies for the growth of the SMEs.
In European Union, the commission emphasizes the participation of
SMEs in the FP6 program but it appears that a lot of time have to
be spent to convince them of the usefulness of their participation
and what are the potential benefits they can have. Moreover they
do not have the means and the ``research culture'' to be the
leader of such a project.

Nowadays we have to find means to face the dilemma of SMEs and VM:
if no ways of dissemination of VM technologies to the SMEs are
found, they will continue to be a mirage. The need for a quick and
appropriate answer is increased by the development of the Virtual
Enterprise  \cite{martinez,mo}. Some propositions, as
Telemanufacturing (e-manufacturing), have appeared
\cite{Offodile}: ``Telemanufacturing is an infrastructure whereby
a firm utilizes services afforded via communications networks and
across information superhighways to perform for the design and
production of items''.
\section*{TRENDS AND EXPLOITABLE RESULTS}

\subsection*{Machine-tool}
The trend in the machine tool manufacturers sector concerning
Virtual Manufacturing is research and development towards the
``Virtual Machine Tool''. One of the goals of the Virtual Machine
Tool is to reduce time and cost for the developing of new machine
tools by introducing virtual prototypes that are characterized by
a comprehensive digital geometrical design, and by the simulation
of (1) the stationary behavior of the machine structure, (2) the
dynamic behavior of moving parts, (3) the changing of signals and
state variables in electronic circuits and (4) by the simulation
of the manufacturing process itself. A second goal is the
simulation of machining processes taking into account the behavior
of the mechanical structure (stresses, deformation,
eigenfrequencies of the moving parts, determination of temperature
distribution, contact behaviour between tool and part, the
determination of cutting forces, vibrations, chatter, ...) and the
mechatronic drives, moving system and the controler.

Nowadays, the simulation activities are isolated from each other.
Current research is combining different types of simulation to
reflect various interdependencies, like e.g. elaborating the
frequency response with FEA and combining this with a bloc
simulation of the machine. The process simulation of forming
processes is almost state-of-the-art in industry, whereas the
simulation of cutting processes is an item for international
research (this excludes the pure NC-program simulation, which is
widely used in industry, but which does not reflect the realistic
behavior of the interaction of machine tool, tool and work piece
during cutting operations).
\subsection*{Automotive}
In the automotive industry, the objective of the Digital Product
Creation Process is to design the product and the manufacturing
process digitally with full visualization and simulation for the
three domains: product, process and resources. In this case a
broad view of manufacturing is adopted and the so-called ``big M''
manufacturing encompasses not only fabrication, assembly and
logistics processes but all functions associated with the entire
{\it product realization process}: from the understanding of
customers' needs to delivery and customer services. A not so
futuristic scenario dealing with automobile can be found in
\cite{bala}.

\begin{figure}[!ht]
  \center
  \epsfig{file=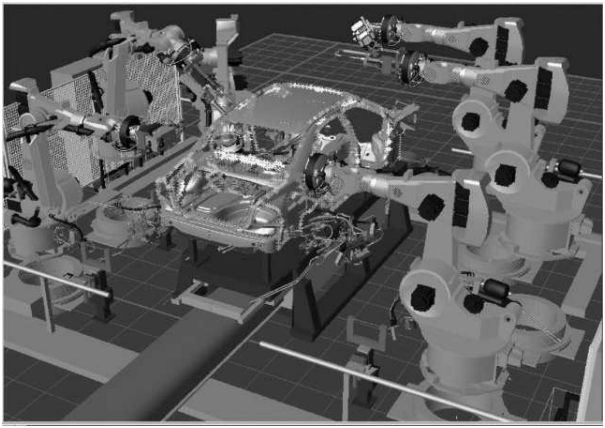, width=8cm}
  \caption{SIMULATION OF A ROBOTIC CELL WITH eM-Workplace (TECNOMATIX)}
  \label{Fig6}
 \end{figure}

The product domain covers the design of individual part of the
vehicle (including all the data throughout the product life
cycle), the process domain covers the detailed planning of the
manufacturing process (from the assignment of resources and
optimization of workflow to process simulation). Flow simulation
of factories and ware houses, 3D-kinematics simulation of
manufacturing systems and robots, simulation of assembly processes
with models of human operators, and FEA of parts of the
automobiles are state-of-the-art (Fig.~\ref{Fig6}).

New trends are focusing on the application of Virtual and
Augmented Reality technologies. Virtual Reality technologies, like
e.g. stereoscopic visualization via CAVE and Powerwall, are
standard in product design. New developments adapt these
technologies to manufacturing issues, like painting with robots.
Developments in Augmented Reality focus on co-operative telework,
where developers located in distributed sites manipulate a virtual
work piece, which is visualized by Head Mounted Displays.
 \begin{figure}[!ht]
  \center
  \epsfig{file=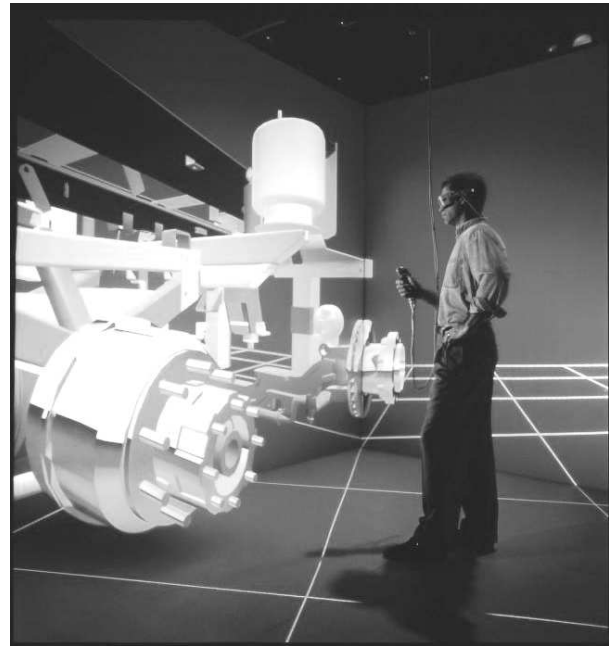, width=8cm}
  \caption{PROJECT REVIEW USING A CAVE (SGI)}
  \label{Fig7}
 \end{figure}
\subsection*{Aerospace}
Virtual Manufacturing in aerospace industry is used in FEA to
design and optimise parts, e.g. reduce the weight of frames by
integral construction, in 3D-kinematics simulation to program
automatic riveting machines and some works dealing with augmented
reality to support complex assembly and service tasks (the worker
sees needed information within his glasses).

\begin{figure}[!ht]
\center
  \epsfig{file=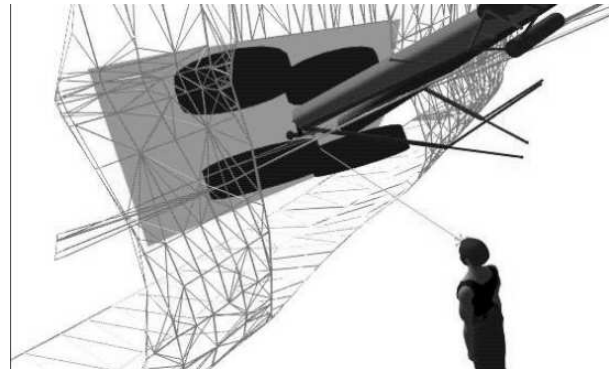, width=8cm}
  \caption{CHECK THE VISUAL ACCESSIBILITY UNDER A TRAP OF AN AIRCRAFT}
  \label{Fig5}
 \end{figure}

The simulation of human tasks with mannequins allows the definition of useful
virtual environnement for assembly, maintenance and training
activities  \cite{chedmail}. The access and visibility task taking into
account ergonomic constraints are defined to check the ability of
mannequin in virtual environments, as depicted in Fig.~\ref{Fig5}.
\subsection*{Drawbacks - Hot topics}
As shown before, a lot of tools and technologies are available in
the field of Virtual Manufacturing and even if  gains they can
offer are well documented in LMEs context their uses are not
prevalent in SMEs. We can propose several explanations. The more
common is the high capital investment (material, software, human)
which is required to set up a VM system. The second explanation
deals with the availability of the simulation models: often at
each level a new model has to be built even if it has already done
before. The third explanation is linked to the VM dependance on
the supporting IT technologies: the compatibility of its software
and hardware is essential for its effectiveness. All the problem
dealing with data exchange, capacity and accessibility of networks
have to be taken into account. In this case one can add the
compatibility between human and computer (human-machine
interface).

Taking these drawbacks as a starting point, several hot topics can
be proposed in the VM research area.

-{\it Integration of simulation systems in planning and design tools:}
Any kind of planning activity can be supported and improved by simulation. The goal to be reached
is to implement simulation systems into planning and design tools so that the planner can
obtain the benefits from the simulation with a minimum of extra effort.

-{\it Automatic generation of simulation models:}
Usually, CAD data has to be manipulated to use it for simulation models. The aim is to
automatically create ready-to-run simulation models out of CAD data with additional information.
Adaptation of a model to specific needs instead of building a totally new one.

-{\it Distributed simulation, optimisation and control:}
In IT research, multi-agent systems get more and more used. This technology can also be applied
to represent distributed systems with autonomous decision making, like it is in a shop floor.
In combination with simulation, multi-agent systems can be used for distributed problem solving and optimisation.
MAS allows the integration of all simulation tools to achieve virtual manufacture of components.

-{\it Hybrid simulation:}
A very hot topic is the combination of real and simulated hardware in machine tool and manufacturing
system development. Real devices, like the machine controller, would be linked to a simulated model
of the machine to test the machine's behaviour during manufacturing.

-{\it Human-computer interfaces:}Users expect to interact with the computer in a human like manner.
We have to develop good interfaces not only graphical but also mixtures of text, voice, visual, ...

-{\it Virtual prototyping:}
The aim is to come to reliable and very precise simulation models, which are able to show nearly realistic
behaviour under static and dynamic stress. The task is to combine results of different kinds of simulation to
predict a nearly realistic behaviour of machine tool, tool and work piece during machining.
\section*{CONCLUSION}
As a conclusion of this paper, we can say that we have now reached
a point where everyone can use VM. It appears that VM will
stimulate the need to design both for manufacturability and
manufacturing efficiency. Nowadays, even if there is still a lot of work
to do, all the pieces are in place for Virtual Manufacturing to
become a standard tool for the design to manufacturing process:
(i) computer technology is widely used and accepted, (ii) the
concept of virtual prototyping is widely accepted, (iii) companies
need faster solutions for cost / time saving, for more accurate
simulations, and (iv) leading companies are already demonstrating
the successful use of virtual manufacturing techniques.
\bibliographystyle{asmems4}
\begin{acknowledgment}
This work has been done thanks to the EC under framework 5
``Thematic network on Manufacturing Technologies'' (MANTYS),
carried out under the ``competitive and sustainable growth''
programme (contract ref. G1RT-CT- 20001-05032)  \cite{depince2}.
\end{acknowledgment}
\def\refname{\large References}
\bibliographystyle{}

\end{document}